\documentclass[sn-chicago]{sn-jnl}

\jyear{2021}%

\theoremstyle{thmstyleone}%
\theoremstyle{thmstyletwo}%

\theoremstyle{thmstylethree}%
\usepackage{subfig}
\usepackage{pifont}
\newcommand{\xmark}{\ding{55}}%
\usepackage{multirow}
\usepackage[super]{nth}
	\usepackage{hyperref} 
\hypersetup{
	colorlinks=true,       
	linkcolor=red,        
	citecolor=blue,        
	filecolor=magenta,     
	urlcolor=black
}

\begin{document}

\title[\hfill]{Anomaly Detection in Additive Manufacturing Processes using Supervised Classification with Imbalanced Sensor Data based on Generative Adversarial Network}


\author[1]{\fnm{Jihoon} \sur{Chung}}\email{\hfill}

\author[2]{\fnm{Bo} \sur{Shen}}\email{\hfill}

\author*[1]{\fnm{Zhenyu (James)} \sur{Kong}}\email{zkong@vt.edu}

\affil[1]{Grado Department of Industrial and Systems Engineering, Virginia Tech, Blacksburg, US}

\affil[2]{Department of Mechanical and Industrial Engineering, NJIT, Newark, US}



\abstract{Supervised classification methods have been widely utilized for the quality assurance of the advanced manufacturing process, such as additive manufacturing (AM) for anomaly (defects) detection. However, since abnormal states (with defects) occur much less frequently than normal ones (without defects) in a manufacturing process, the number of sensor data samples collected from a normal state is usually much more than that from an abnormal state. This issue causes imbalanced training data for classification analysis, thus deteriorating the performance of detecting abnormal states in the process. It is beneficial to generate effective artificial sample data for the abnormal states to make a more balanced training set. To achieve this goal, this paper proposes a novel data augmentation method based on a generative adversarial network (GAN) using additive manufacturing process image sensor data. The novelty of our approach is that a standard GAN and classifier are jointly optimized with techniques to stabilize the learning process of standard GAN. The diverse and high-quality generated samples provide balanced training data to the classifier. The iterative optimization between GAN and classifier provides the high-performance classifier. The effectiveness of the proposed method is validated by both open-source data and real-world case studies in polymer and metal AM processes.}




\maketitle
\section{Introduction}\label{sec1}
\begin{figure}[!htp]
    \centering
    {\includegraphics[width=0.5\textwidth]{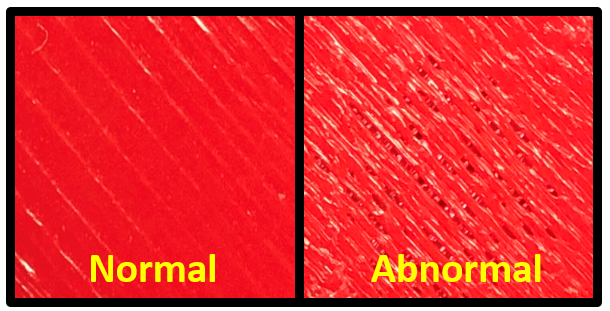}}\vspace{-0.0cm}
    \caption{Normal surface and Abnormal defect in AM process.}
    \label{fig:gan_1}
\end{figure}

Advanced manufacturing processes, such as additive manufacturing (AM), are widely applied in various fields, including aerospace, healthcare, and the automotive industry \citep{liu2019image}. However, a major barrier preventing broader industrial adoption of the processes is the quality assurance of products. For example, surface defects exist, such as under-fill from the Fused Filament Fabrication (FFF) process shown in Fig.~\ref{fig:gan_1}. It is due to highly complex interactions in consecutive layers during printing. The defect causes a deficiency in the mechanical properties of the final product, such as density, tensile strength, and compressive strength \citep{hajalfadul2021building}. Therefore, timely detection of the anomaly in the process is necessary so that corrective actions can be taken to mitigate the defects and improve the quality of products \citep{makes2017standardization}.

Recently, the development of sensor technologies and computational power offered online process data, providing excellent opportunities to achieve effective quality assurance through a data-driven approach \citep{liu2021integrated}. Specifically, the sensor data are usually in high volume in terms of both dimensionality and sampling frequency, having plenty of information about the manufacturing processes \citep{zhou2020pd}. Therefore, it is very beneficial to investigate the underlying relationships between the sensor data and process quality state based on a data-driven perspective \citep{li2021augmented}. Utilizing the sensor data, supervised classification methods have been widely applied to online anomaly detection in the manufacturing process \citep{banadaki2020toward, jin2019autonomous, guo2019fault}. This is because these methods can fully utilize the label information of the process state, resulting in more accurate and reliable detection results than unsupervised methods. \cite{banadaki2020toward, jin2019autonomous, guo2019fault} utilized balanced training data collected from both normal and abnormal states in the manufacturing processes to achieve a high anomaly detection rate from classifiers \citep{choi2021imbalanced}. 
\vspace{-0.5cm}\vspace{-0.0cm}

However, the manufacturing process is usually in a normal state \citep{li2021augmented}. Therefore, balanced training data assumed by the existing work \citep{banadaki2020toward, jin2019autonomous} is expensive and not realistic. In reality, abnormal conditions, such as surface defects in AM process (Fig.~\ref{fig:gan_1}), may happen but rarely. Consequently, the sensor data collected under abnormal states are smaller than those collected in a normal state, and may not be sufficient for training supervised classification methods. It causes imbalanced training data among process states \citep{chawla2004special}, leading to compromised anomaly detection performance in actual manufacturing processes \citep{li2021augmented}.  Specifically, when the number of training data samples from a normal state is significantly more than that of abnormal states, the prediction in classification models tends to be biased towards the normal state (i.e., majority class) \citep{choi2021imbalanced}. This leads to a high probability of misclassifying samples from the abnormal states (i.e., minority classes). 

To overcome this severe issue caused by imbalanced data problems in the manufacturing process, data augmentation can be applied to obtain balanced training data among process states. Simple data augmentation techniques such as rotation, flipping,  and synthetic minority oversampling technique (SMOTE) \citep{chawla2002smote} are widely used for balancing training data in the supervised classification method for the manufacturing process because of their simplicity in application \citep{cui2020metal,lee2020automated,mycroft2020data}. However, such methods consider only local information; therefore, they cannot reflect the entire data distribution and overcome the problem of overfitting \citep{douzas2018effective, mikolajczyk2018data}. As a result, these methods are unsuitable for generating realistic and diverse data on abnormal states in manufacturing applications \citep{fathy2020learning,ranasinghe2019generating}. Recently, the generative adversarial network (GAN) \citep{goodfellow2014generative} has been actively used for data augmentation in manufacturing. This is because GAN generates more representative data than the simple augmentation method by learning the entire distribution of actual data through two neural networks: discriminator and generator \citep{antoniou2017data,kiyasseh2020plethaugment}. For example, \cite{gobert2019conditional} used conditional GAN to generate layer images from the metal AM process. In addition, \cite{li2021augmented} developed a novel GAN method considering temporal orders of the sensor signal from the polymer AM process to generate the balanced training signal data.

Utilizing GAN, the existing studies achieved high classification performance by training the classifier with the generated balanced training samples. The generated samples of each process state are realistic by learning the distribution of actual data. However, features enabling differentiation between process states (i.e., state-distinguishable features) that can further improve the classification performance have not been considered in the generation process from GAN in the previous studies \citep{gobert2019conditional,li2021augmented}.

This paper proposes a novel GAN-based data augmentation method in manufacturing to generate realistic and state-distinguishable generated data. Compared to a standard GAN consisting of two players, i.e., discriminator and generator, the proposed method comprises three players, including a classifier. All players are jointly optimized in the proposed approach to meet their respective goals. Specifically, the generator learns to generate data deceiving the discriminator. In contrast, a discriminator learns to distinguish whether data is from a generator or an actual process. This adversarial learning results in generating realistic samples from the generator. At the same time, the generator and classifier cooperate to generate distinctive samples among process states in the manufacturing process. Specifically, the classifier guides the generator to create samples that could benefit classification results. Then, the classifier is trained with balanced training data supplemented from generated samples of abnormal states. This iterative learning among three-player finally provides the classifier with high performance. 

Furthermore, the proposed method provides two terms in the objective function of the discriminator to improve the training stability of standard GAN. First, the proposed method regularizes the gradient of the discriminator \citep{fedus2017many}. Second, the proposed approach provides an additional task to the discriminator compared to standard GAN, preventing the discriminator from discerning the origin of data very well \citep{huang2021enhanced}. Both terms improve the training stability by preventing gradient exploding in training the generator. The contributions of this work are summarized as follows: 
\begin{itemize}
    	\item  From the methodological point of view, this paper proposes a novel data augmentation method that standard GAN and classifier are jointly optimized with techniques to stabilize the learning process of standard GAN. The  generated samples from the generator provide balanced training data to the classifier. The iterative optimization between GAN and classifier provides a classifier with superior performance. 
    	\item 	From the application perspective, the proposed method is applied to anomaly detection in actual additive manufacturing processes. The technique successfully detects process anomalies when highly imbalanced data sets exist. The effectiveness of the proposed method is validated from polymer and metal AM processes, namely, Fused Filament Fabrication (FFF) and Electron Beam Melting (EBM) processes.
\end{itemize}

The rest of this paper is organized as follows. A brief review of related research work is provided in Section~\ref{s:sec.2}. The proposed methodology is presented in Section~\ref{s:sec.3}, followed by case studies with open-source and actual AM data sets to validate its effectiveness in Section~\ref{s:sec.4}. Finally, conclusions are discussed in Section~\ref{s:sec5}.

\section{Review of Related Work} \label{s:sec.2}
The relevant existing studies on anomaly detection approaches for the manufacturing processes are briefly reviewed first in Section~\ref{s:sec2.1}, and subsequently, the existing data augmentation methods are summarized in Section~\ref{s:sec2.2}. Afterward, the research gaps in the current work are identified in Section~\ref{s:sec2.3}.
\subsection{Anomaly Detection Approaches for the Manufacturing Processes} \label{s:sec2.1}
The anomaly detection approaches for manufacturing processes have been extensively studied. Recently, deep learning-based methods have been widely used. For example, \cite{kyeong2018classification} proposed using convolutional neural networks (CNNs) to classify mixed-type defect patterns in wafer bin maps in the semiconductor manufacturing process. \cite{kwon2020deep} used deep neural networks to accurately classify melt pool images in the metal AM process concerning different laser powers, resulting in different porosity levels. \cite{zhang2020new} converted the vibration signal from the drive end bearing to the images. The images were used as training data for CNN for bearing fault diagnosis. In addition, \cite{jia2019rotating}  applied CNN to extract the features from the infrared thermal images for the fault diagnosis of rotating machinery in the manufacturing process.  

In addition to the deep learning-based methods, various machine learning-based methods are utilized in anomaly detection in the manufacturing processes. For example, \cite{montazeri2018sensor} proposed a graph-theoretic approach to differentiate the distinctive thermal signatures of melt pool images, leading to poor abnormal surface finish from the metal AM process. \cite{mahmoudi2019layerwise} also developed a novel anomaly detection framework in the metal AM process. The approach classified the process state by accounting for the spatial dependence among successive melt pools through the Gaussian process. For the polymer AM process, \cite{liu2021integrated} proposed a feature extraction method based on manifold learning to diagnose surface defects such as under-fill. \cite{shen2020clustered} also proposed a novel supervised feature extraction method to extract discriminant and informative features from the surface states in the FFF process. In addition to the several AM processes, \cite{bhat2016tool} presented a classification method to diagnose the cutting tool condition in the manufacturing process by analyzing the machined surface. Specifically, a kernel-based support vector machine classifier was trained with the features extracted from the gray-level co-occurrence matrix of machined surface images. \cite{wang2019wafer} proposed weight masks to extract rotation-invariant features for wafer map failure pattern detection in the semiconductor manufacturing process. 

\subsection{Data Augmentation Methods} \label{s:sec2.2}
The existing data augmentation methods from the literature can be divided into two categories, sampling-based approaches, e.g., an oversampling technique, and deep learning-based approaches, e.g., GAN. 

Sampling-based approaches such as oversampling methods generate synthetic samples based on the existing samples, including interpolation, to balance the data. The most popular oversampling method is the synthetic minority oversampling technique (SMOTE) proposed by \cite{chawla2002smote}. It also has several extensions \citep{fernandez2018smote}. For example, many rules for regions of data to be oversampled are settled down, such as Borderline-SMOTE (B-SMOTE) \citep{han2005borderline} and the adaptive synthetic sampling approach (ADASYN) \citep{he2008adasyn}. Furthermore, there exist hybrid methods, such as SMOTE editing the nearest neighbor (SMOTENN) \citep{batista2004study} and SMOTE-Iterative partitioning filter (SMOTE-IPF) \citep{saez2015smote}. They remove unsuitable samples after SMOTE-based oversampling.

Recently, deep learning has become popular in data augmentation due to its  powerful computation for high-dimensional data \citep{shorten2019survey}. GAN is one of the most widely applied approaches, which was first introduced by \cite{goodfellow2014generative}. GAN-based work for data augmentation exploits the GAN models, such as deep convolutional GAN (DCGAN) \citep{wang2017cgan}, cycle GAN \citep{zhang2018deep}, and conditional GAN (cGAN) \citep{douzas2018effective} to supplement the limited actual data by generated data. Balancing GAN (BAGAN) \citep{mariani2018bagan} is a modified version of an auxiliary classifier GAN \citep{odena2017conditional} that focuses on the generation of minority class samples. Moreover, \cite{huang2021enhanced} proposed an enhanced version of BAGAN (BAGAN-GP) that overcomes the unstable training issue in BAGAN by providing an improved initialization method and gradient penalty technique \citep{gulrajani2017improved}. Recently, \cite{choi2021imbalanced} proposed  a three-player GAN that consists of a discriminator, generator, and classifier. In this method, the processes of generating samples and training classifiers are jointly optimized. It generates both realistic and state-distinguishable generated samples that are beneficial to improving the classification results. However, this work follows a standard GAN structure \citep{goodfellow2014generative}  that is prone to unstable learning resulting in limited diversity and poor quality of generated samples \citep{zhu2019can}.

\subsection{Research Gap Analysis} \label{s:sec2.3}
The studies summarized in Section~\ref{s:sec2.1} contributed to anomaly detection in the manufacturing process. However, the work does not consider the imbalanced training data that usually occurs in reality. Instead, the literature requires sufficient data from abnormal states, which is highly time-consuming and prohibitively costly. Research efforts in Section~\ref{s:sec2.2} provided data augmentation work to overcome the imbalanced training data issue in the manufacturing process. However, the sampling-based methods consider only local information; therefore, they cannot reflect the entire data distribution \citep{douzas2018effective}. Recently, GAN-based methods have been widely used because of their superior performance in data augmentation. However, the process of generating samples through GAN and the process of training a classifier with the generated samples are handled independently in most of the studies. It is critical to generate realistic samples that follow the distribution of actual samples through GAN. However, it is also important that the generated samples help improve the classifier performance as training data. To achieve this, the realistic samples must be well-distinguishable among the process states. Thus, this paper proposes a novel GAN-based data augmentation method for additive manufacturing applications where the processes of generating samples and training a classifier are jointly optimized. Furthermore, the proposed method regularizes the gradient of the discriminator and provides an additional task to the discriminator compared to standard GAN. This improves the training stability that prevents gradient exploding of the generator. It results in a better quality of generated samples, improving the classifier's performance compared to the existing work.
The effectiveness of the proposed method is validated with imbalanced surface images from actual AM processes described in Section~\ref{s:sec.4}. 

\section{Proposed Research Framework} \label{s:sec.3}
This section proposes a novel GAN-based data augmentation method. The structure of the proposed method is illustrated in Section~\ref{s:sec3.1}. Specifically, the objective function of the proposed method is described in Section~\ref{s:sec3.2}, followed by the training procedure in Section~\ref{s:sec3.3}.
\subsection{Three-Player Structure for Imbalanced Data Learning} \label{s:sec3.1}
\begin{figure}[!htp]
    \centering
    {\includegraphics[width=1.0\textwidth]{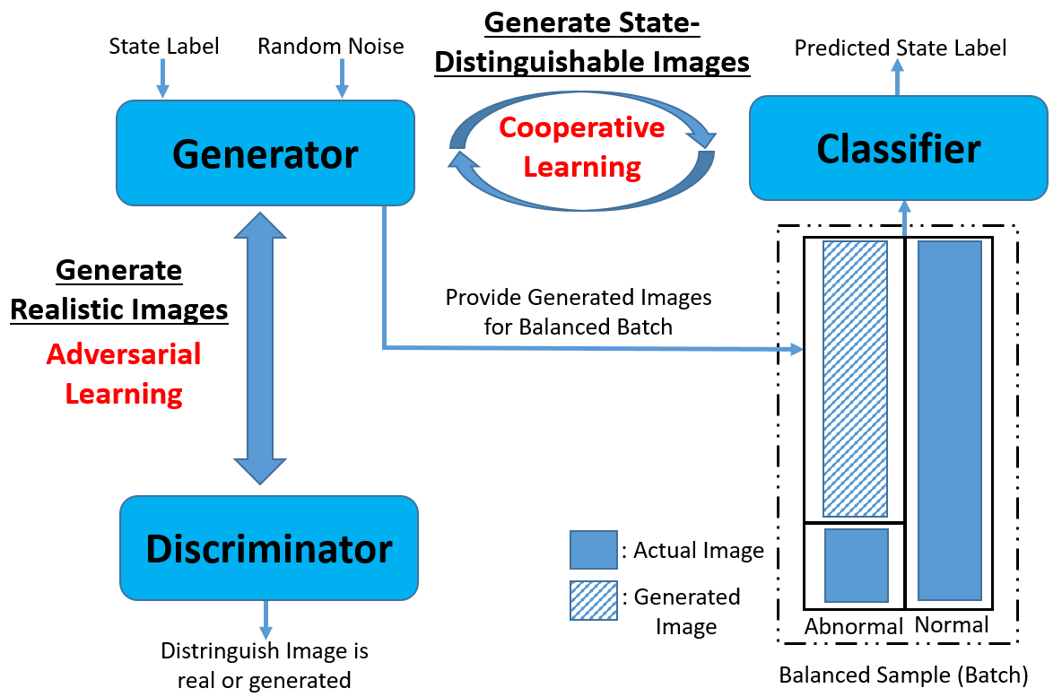}}\vspace{-0.0cm}
    \caption{Structure of the proposed method.}
    \label{fig:gan_2}
\end{figure}

Fig.~\ref{fig:gan_2} shows the structure of the proposed method, which consists of  three players: a discriminator, a generator, and a classifier.
\vspace{-0.0cm}
The generator generates images of the manufacturing process from the random noise and corresponding state label. Among the generated images, the images of abnormal states are combined with the actual imbalanced manufacturing process images, providing a balanced training sample to the classifier. The proposed method offers adversarial and cooperative learning to make the generated images beneficial to the classifier's performance. The roles of these two are provided as follows.
\begin{itemize}
    \item Adversarial learning: The relationship between discriminator and generator follows the adversarial relationship from the GAN structure. The relationship enables both networks to compete with each other, resulting in realistic generated samples of the manufacturing process from the generator.
    \item Cooperative learning: The cooperative relationship between the generator, and the classifier enables the generator to generate samples that are distinguishable among process states in the manufacturing process (i.e., state-distinguishable samples) from the classifier. 
\end{itemize}
Based on these two relationships, the generator generates samples of abnormal states in the manufacturing process with both properties (i.e., realistic and state-distinguishable). The generated samples are added to actual ones and provided as a balanced training batch that the number of training samples passing through the network at one time of a classifier. 
The iterative learning process among these three players finally provides a classifier with high performance. The detailed objective function and training procedures are explained in the following sections.

\subsection{Objective Functions for Three-Player} \label{s:sec3.2}
In this subsection, the review of the generative adversarial network in the manufacturing process is described in Section~\ref{subsub:3.2.1} initially. Then, the objective functions of the discriminator, generator, and classifier are illustrated in Sections~\ref{subsub:3.2.2},~\ref{subsub:3.2.3}, and~\ref{subsub:3.2.4}, respectively.  
\subsubsection{Generative Adversarial Network in the Manufacturing Process}\label{subsub:3.2.1}
The idea of a Generative Adversarial Network (GAN) is to train two networks, namely, generator $G$ and discriminator $D$, with a minimax game for $V(D,G)$ demonstrated in Eq.~\eqref{eq:gan_1} \citep{wang2017cgan}.
\begin{equation}\label{eq:gan_1}
\begin{aligned}
   \min_{G}\max_{D}  &V(D,G) =  \mathbb{E}_{x_{a}\sim P(X_{a})}[\text{log}(D(x_{a}))] \\
  &  +\mathbb{E}_{z \sim P({Z})}[\text{log}(1-D(G(z))],
\end{aligned}
\end{equation}
where $z$ is the random noise, and $x_{a}$ denotes actual samples from the manufacturing process. Specifically, the generator is to generate samples of the manufacturing process $G(z)$ from $z$, and the discriminator is to distinguish whether the origin of input samples is from actual ($x_{a}$) or generator ($G(z)$). In other words, the discriminator is to discern the input samples, while the generator synthesizes artificial samples to deceive the discriminator.
This adversarial learning results in the distribution of newly generated samples, close to the underlying distribution of the actual samples in the manufacturing process, $P(X_{a})$.

As shown in Fig.~\ref{fig:gan_2}, the proposed method needs to provide balanced training samples among the manufacturing process states in every batch of a classifier. 
To achieve this, conditional GAN \citep{douzas2018effective} is applied in the proposed method making the process state label attached to the input of the generator and discriminator. It enables users to determine the number of generated samples from abnormal states of the manufacturing process to make a balanced batch. The objective function of conditional GAN is demonstrated in Eq.~\eqref{eq:gan_2} as follows. 
\begin{equation}\label{eq:gan_2}
\begin{aligned}
      \min_{G}\max_{D}&V(D,G)=\mathbb{E}_{(x_{a},y_{a})\sim P(X_{a},Y_{a})}[\text{log}(D(x_{a},y_{a}))] \\
      &+\mathbb{E}_{(z,y_{g})\sim P(Z,Y_{g})}[\text{log}(1-D(G(z,y_{g}),y_{g})],
\end{aligned}
\end{equation}
where $y_{a}$ and $y_{g}$ denote the process state label of an actual and generated sample, respectively.
\subsubsection{Objective Function of Discriminator}\label{subsub:3.2.2}
The objective of the discriminator in the proposed method is to maximize Eq.~\eqref{eq:gan_2} through adversarial learning with the generator. Specifically, the discriminator learns to distinguish the input ($x_{a},y_{a}$) and ($G(z,y_{g}),y_{g}$) are actual and generated, respectively. In addition, the proposed method provides two additional terms for stabilizing the learning process, as the training of the GAN is unstable and challenging to converge due to the gradient exploding issue in adversarial learning \citep{tao2020alleviation, arjovsky2017towards}.

First, the proposed method regularizes the gradient of the discriminator by providing the gradient penalty in the form of Eq.~\eqref{eq:gan_3} as follows. 
\begin{equation}\label{eq:gan_3}
\mathbb{E}_{(\hat{x},y_{a})\sim P(\hat{X},Y_{a})}[(\lVert \nabla_{(\hat{x},y_{a})}D(\hat{x},y_{a}) \rVert_{2} -1)^{2}],
\end{equation}
where $\hat{x}=\alpha x_{a} +(1-\alpha)G(z)$, and $\alpha$ is sampled uniformly between 0 and 1. 
It enforces 1-Lipschitz continuity to the discriminator \citep{davenport1951principle}. The term is widely used in the previous work \citep{gulrajani2017improved,fedus2017many,huang2021enhanced} to overcome the limited diversity and poor quality of generated samples from GAN caused by unstable adversarial learning. 

Second, the proposed method provides an additional input consisting of the actual sample ($x_{a}$) and a mislabeled process state ($y_{m}$) to the discriminator. The discriminator learns to distinguish that the input ($x_{a},y_{m}$) is not an actual but generated sample because of the mislabeled process state by minimizing the following quantity. 
\begin{equation*}
    -\mathbb{E}_{(x_{a},y_{m})\sim P(X_{a},Y_{m})}[\text{log}(1-D(x_{a},y_{m})],
\end{equation*}
where $y_{m}$ is randomly sampled from the remaining labels in the manufacturing process  except for the actual state label. Since the input provides  an additional task for the discriminator to learn, it prevents the discriminator distinguishes very well between actual and generated samples before the generator approximates the actual sample distribution of the manufacturing process. Otherwise, it causes unstable learning of GAN through exploding or vanishing the gradient of the generator \citep{tran2018dist, arjovsky2017towards}.

In summary, the objective function of the discriminator ($L^{D}$) in the proposed method is to minimize the
following equation consisting of several losses.
\begin{equation}\label{eq:gan_5}
    \begin{aligned}
  L^{D}(&Z, X_{a},  Y_{a},  Y_{g}, Y_{m})= \\
  &\underbrace{-\mathbb{E}_{(x_{a},y_{a})\sim P(X_{a},Y_{a})}[\text{log}(D(x_{a},y_{a}))]}_{\text{loss from actual sample in discriminator}} \\
    & \underbrace{-\mathbb{E}_{(z,y_{g})\sim P(Z,Y_{g})}[\text{log}(1-D(G(z,y_{g}),y_{g})]}_{\text{loss from generated sample in discriminator}} \\
    & \underbrace{-\mathbb{E}_{(x_{a},y_{m})\sim P(X_{a},Y_{m})}[\text{log}(1-D(x_{a},y_{m})]}_{\text{loss from mislabeled sample in discriminator}}  \\ 
    & + \lambda\underbrace{\mathbb{E}_{(\hat{x},y_{a})\sim P(\hat{X},Y_{a})}[(\lVert \nabla_{(\hat{x},y_{a})}D(\hat{x},y_{a}) \rVert_{2} -1)^{2}]}_{\text{loss from gradient penalty}},
\end{aligned}
\end{equation}
where $\lambda$ is the coefficient of the gradient penalty term. The first three losses in Eq.~\eqref{eq:gan_5} are related to losses when the discriminator misclassified the origin of the actual, generated, and mislabeled sample. The last loss represents the loss related to the gradient of the discriminator.
\subsubsection{Objective Function of Generator}\label{subsub:3.2.3}
The objective of the generator is to generate samples via learning the distribution of actual samples of the manufacturing process ($P(X_{a})$) by minimizing Eq.~\eqref{eq:gan_2}. Alternatively, the proposed method trains the generator to maximize Eq.~\eqref{eq:gan_6} to avoid the saturation problem occurring when minimizing Eq.~\eqref{eq:gan_2} in practice \citep{goodfellow2014generative}. 
\begin{equation}\label{eq:gan_6}
 \mathbb{E}_{(z,y_{g})\sim P(Z,Y_{g})}[\text{log}(D(G(z,y_{g}),y_{g})].
\end{equation}

In addition to Eq.~\eqref{eq:gan_6}, the generator in the proposed method has an additional term in the objective function related to the classifier. In contrast to an adversarial relationship with a discriminator, the cooperative relationship is designed between the generator and classifier to generate well-distinguishable samples among process states in the manufacturing process. To achieve this, the classification loss from labels of generated samples is provided to the objective function of the generator (\nth{2} term in Eq.~\eqref{eq:gan_7}). Finally, the generator in the proposed method is trained by minimizing its objective function ($L^{G}$), Eq.~\eqref{eq:gan_7}.
\begin{equation}\label{eq:gan_7}
\begin{aligned}
      L^{G}(Z, Y_{g}) = & \underbrace{-\mathbb{E}_{(z,y_{g})\sim P(Z,Y_{g})}[\text{log}(D(G(z,y_{g}),y_{g}))]}_{\text{loss from generated sample in discriminator}}\\
      &\underbrace{-\mathbb{E}_{(z,y_{g})\sim P(Z,Y_{g})}[ y_{g}\text{log}(C(G(z,y_{g})))]}_{\text{loss from generated sample in classifier}},
\end{aligned}
\end{equation}
where $-\mathbb{E}_{(z,y_{g})\sim P(Z,Y_{g})}[ y_{g}\text{log}(C(G(z,y_{g})))]$ (\nth{2} term in Eq.~\eqref{eq:gan_7}) denotes the cross-entropy loss of generated samples from the classifier.
\subsubsection{Objective Function of Classifier}\label{subsub:3.2.4}
 The objective function of the classifier ($L^{C}$) consists of the classification loss from both actual and generated samples of the manufacturing process as Eq.~\eqref{eq:gan_8}. As described in Fig.~\ref{fig:gan_2}, the samples from the generator are supplemented with actual samples from the manufacturing process to make  balanced training data in every batch of the classifier. The classifier is optimized to minimize the classification loss from both actual and generated sample by minimizing Eq.~\eqref{eq:gan_8}.  
\begin{equation}\label{eq:gan_8}
\begin{aligned}
    L^{C}(Z, X_{a},&Y_{a}, Y_{g}) =  \underbrace{-\mathbb{E}_{(x_{a},y_{a}) \sim P(X_{a},Y_{a})}[y_{a}\text{log}(C(x_{a}))]}_{\text{loss from actual sample in classifier}}\\
    &\underbrace{-\mathbb{E}_{(z,y_{g})\sim P(Z, Y_{g})}[y_{g}\text{log}(C(G(z,y_{g}))]}_{\text{loss from generated sample in classifier}}.      
\end{aligned}
\end{equation}
In particular, $-\mathbb{E}_{(z,y_{g})\sim P(Z, Y_{g})}[y_{g}\text{log}(C(G(z,y_{g}))]$, a common term in both Eqs.~\eqref{eq:gan_7} and~\eqref{eq:gan_8} enables cooperative learning between the generator and classifier.

\vspace{-0.0cm}\subsection{Training Procedure} \label{s:sec3.3}
To train the three players in the proposed method, the method adopts an alternating gradient descent method among the training of generator, discriminator, and classifier. 
Before starting alternating optimization, the auto-encoder is pre-trained with the existing imbalanced samples from the manufacturing process. Auto-encoder is trained to minimize the reconstruction error of inputs. Auto-encoder is widely used for the initialization of the generator in GAN because it leads to stable learning and helps the generator learn common data set knowledge \citep{mariani2018bagan}. 
The pre-trained decoder from an auto-encoder is initialized as the generator in the proposed method.
After the pre-training step, the three players are optimized alternatively. 
First, the discriminator is trained with a batch from actual and generated samples to minimize Eq.~\eqref{eq:gan_5}. Sequentially, a batch from generated samples is utilized to update the generator by minimizing Eq~\eqref{eq:gan_7}. Finally, the classifier is trained by minimizing Eq.~\eqref{eq:gan_8} with balanced training data from all the process states in the manufacturing process. Specifically, a batch ($m$) from actual data is sampled first. Then, the remaining samples from abnormal states ($m_g$) are generated from the generator to make a balanced training set.
The alternating training procedure is iterated until it reaches the pre-defined epochs. The overall training procedure of the proposed method is illustrated in Algorithm~\ref{alg:alg1}. The parameters of a discriminator, generator, and classifier are denoted as $\theta_{d}, \theta_{g},$ and $\theta_{c}$, respectively.
\vspace{-0.5cm}\begin{algorithm}[!htb]
\begin{algorithmic}
    \Require Initialize the parameters of three players (i.e., $\theta_{d}$, $\theta_{g}$, and $\theta_{c}$). \vspace{-0.0in}\\ \hspace{0.47in} $\text{P-Epoch}$: Number of epochs in the pre-training.\vspace{-0.0in}\\ \hspace{0.47in}
$\text{A-Epoch}$: Number of epochs in the alternating loop.\vspace{-0.0in}\\ \hspace{0.47in}
$\text{Batch size}$: Size of samples ($m$) in each batch.\\ \vspace{-0.0in}
\hspace{-0.1in}\textbf{Procedure:}
\vspace{-0.0in}\\ \hspace{0.48in}[\textbf{Pre-Training Generator}] 
\vspace{-0.0in}\\ \hspace{0.70in} Initialize: epoch=1  
\vspace{-0.0in}\\ \hspace{0.70in} \textbf{while} epoch $\leq$ \text{P-Epoch} \textbf{do}  
\vspace{-0.0in}\\ \hspace{0.83in} Sample a batch of $m$ samples from $X_{a}$
\vspace{-0.0in}\\ \hspace{0.83in} Train  auto-encoder ($\theta_{g}$) 
\vspace{-0.0in}\\ \hspace{0.83in} $\text{epoch + +}$  
\vspace{-0.0in}\\ \hspace{0.72in} \textbf{end while} 
\vspace{-0.0in}\\ \hspace{0.55in}[\textbf{Alternating Optimization}] 
\vspace{-0.0in}\\ \hspace{0.72in} Initialize: epoch=1  
\vspace{-0.0in}\\ \hspace{0.72in} \textbf{while} epoch $\leq$ \text{A-Epoch} \textbf{do}  
\vspace{-0.0in}\\ \hspace{0.85in}  Sample $m$ samples from  $X_{a},Z,Y_{g},Y_{m}$      
\vspace{-0.0in}\\ \hspace{0.85in} \textbf{Train a discriminator $(\theta_{d})$ by minimizing Eq.~\eqref{eq:gan_5}} 
\vspace{-0.0in}\\ \hspace{0.85in} Sample $m$ samples from  $Z,Y_{g}$   
\vspace{-0.0in}\\ \hspace{0.85in} \textbf{Train a generator $(\theta_{g})$ by minimizing Eq.~\eqref{eq:gan_7}}   
\vspace{-0.0in}\\ \hspace{0.85in} Sample $m$ samples from $X_{a}$  
\vspace{-0.0in}\\ \hspace{0.86in} Sample $m_{g}$  from $Z,Y_{g}$ of abnormal states to balance the batch  
\vspace{-0.0in} \hspace{0.86in} \textbf{Train a classifier $(\theta_{c})$ by minimizing Eq.~\eqref{eq:gan_8}} 
\vspace{-0.0in}\\ \hspace{0.85in} $\text{epoch + +}$
\vspace{-0.0in}\\ \hspace{0.75in} \textbf{end while}
    \caption{Training Procedures of the proposed method.}\label{alg:alg1}
\end{algorithmic}
\end{algorithm}\vspace{-0.5cm}
\section{Real-World Case Studies}\label{s:sec.4}
Several real-world case studies are provided to show the effectiveness of the proposed method.
Section~\ref{s:sec4.1} demonstrates the advantages of the proposed method based on the ablation study. The method is deeply self-analyzed in various aspects.  Sections~\ref{s:sec4.2},~\ref{s:sec4.3}, and~\ref{s:sec4.4} provide comparative case studies with benchmark methods across multiple data sets. Specifically, open-source data set is used in Section~\ref{s:sec4.2}. In addition, the surface image data from two actual AM processes, namely, Fused Filament Fabrication (FFF) and Electron Beam Melting (EBM) processes, are utilized in Sections~\ref{s:sec4.3} and~\ref{s:sec4.4}, respectively, to demonstrate the effectiveness of the proposed method in an actual AM process. The performance is evaluated by the classification results from the imbalanced training data set. The framework of all case studies is Keras with TensorFlow backend. The studies are performed by an NVIDIA Tesla P4 GPU with 8GB memory.

\vspace{-0.0cm}\textbf{1) Benchmark Methods}: For the benchmark methods, both the sampling-based and GAN-based approaches described in Section~\ref{s:sec2.2} are used. In the sampling-based methods, SMOTE \citep{chawla2002smote}, B-SMOTE \citep{han2005borderline}, and ADASYN \citep{he2008adasyn}, which are implemented in the imbalanced-learn library in python, are adopted. For the GAN-based approaches, two state-of-the-art methods, CDRAGAN \citep{huang2021enhanced}, and BAGAN-GP \citep{huang2021enhanced} are selected. In addition, Cooperative GAN \citep{choi2021imbalanced}, which jointly optimizes GAN and the classifier, is determined as one of the benchmark methods. Finally, classification performance without any data augmentation method is provided as the baseline.

\vspace{-0.0cm}\textbf{2) Hyperparameters and Experimental Settings}: The optimizer for the proposed method is the Adam algorithm \citep{kingma2014adam} with a learning rate of 0.0002 and momentums of 0.5 and 0.9 \citep{huang2021enhanced}. To make the networks in the proposed method irrespective of the image size, all the image inputs are resized as 64$\times$64$\times$channels. The dimension of the random noise ($z$) is  128. Since the proposed method deals with image data, the auto-encoder with convolution layers is used for pre-training \citep{huang2021enhanced}. In addition, the discriminator and generator are designed with a convolution layer, leakyReLu layer, and transpose convolution layer. For the classifier, Convolutional Neural Network (CNN) is utilized in the case studies since CNN extracts the features from the raw data directly, resulting in superior performance in image classification \citep{dhillon2020convolutional}. For a fair comparison, the CNN with the same hyperparameters is used for all the methods. Besides, the unique hyperparameters of each method, such as the scheduling parameter in the Cooperative GAN \citep{choi2021imbalanced}, are searched within a specific range following the guidelines provided in the literature and determined with the values that showed the best performance. The detailed hyperparameters of the generator, discriminator, and classifier are provided in Appendix~\ref{app:app1}.

\vspace{-0.0cm}\textbf{3) Performance Evaluation Measure}: The classification performance is evaluated by the value of the F-score, Precision,  and Recall \citep{powers2020evaluation}. Precision and Recall are related to the level of type I and type II errors, respectively. F-score can be formulated by Eq.~\eqref{eq:gan_9}, which is the combination of Precision and Recall. 
\begin{equation}\label{eq:gan_9}
    \text{F-score} = 2\times \frac{\text{Precision} \times \text{Recall}}{\text{Precision}+\text{Recall}}.
\end{equation}

Since this paper aims to improve the classification accuracy with the imbalanced training data, case studies with various balanced ratios  are provided. A balanced ratio is the ratio between the training data size of the minority and the majority classes. All the case studies were repeated ten times, and the average of ten repetitions was provided as the performance measure.
\vspace{-0.0cm}\subsection{Ablation Studies} \label{s:sec4.1}
The ablation study is conducted with MNIST fashion data \citep{xiao2017fashion}. From the 1000 images of each of three labels, namely, T-shirt, Pullover, and Dress, imbalanced training data is constructed as described in  Table~\ref{tab:gan_table1}. To provide the challenging task, the balanced ratio between the majority and minority classes is 0.10. The remaining images are used as testing data.  
\vspace{-0.6cm}\begin{table}[!htb]
\renewcommand{\arraystretch}{1.2}
\centering
\caption{Imbalanced training data samples in ablation studies.}
\label{tab:gan_table1}
\resizebox*{1.0\textwidth}{!}{%
\begin{tabular}{cccccc}
\hline \hline
 \textbf{\begin{tabular}[c]{@{}c@{}}\vspace{-0.0cm}Data\\Set\end{tabular}}        &  \textbf{\begin{tabular}[c]{@{}c@{}}\vspace{-0.0cm}Majority\\Class\end{tabular}} & \textbf{\begin{tabular}[c]{@{}c@{}}\vspace{-0.0cm}Minority\\Class\end{tabular}}                     & \textbf{\begin{tabular}[c]{@{}c@{}}\vspace{-0.0cm}Balanced\\Ratio\end{tabular}}  & \textbf{\begin{tabular}[c]{@{}c@{}}\vspace{-0.0cm}Majority Class\\Training Samples\end{tabular}}& \textbf{\begin{tabular}[c]{@{}c@{}}\vspace{-0.0cm}Minority Class\\Training Samples\end{tabular}} \\  \hline
 
MNIST\ fashion & T-shirt  & Pullover, Dress & 0.10  & 800        &80\\                                          
\hline \hline                               
\end{tabular}
 }
\end{table}\vspace{-0.5cm}

The study is carried out by sequentially adding each ablation component since each component cannot be implemented without the previous components. The role of each component is validated through the ablation of a baseline and three variants, as illustrated in Table~\ref{tab:gan_table2}. Baseline trains classifier with imbalanced training data without any data augmentation. Instead, Variant 1 uses a conditional version of standard GAN (i.e., CGAN \citep{mirza2014conditional}) to provide balanced training data of the classifier by generating additional minority class samples ($G(z, y_{g})$ in Section~\ref{s:sec.3}). Furthermore, Variant 2 jointly trains the classifier and CGAN by providing cross-entropy losses ($y_{g}\text{log}(C(G(z,y_{g})))$, and  $y_{a}\text{log}(C(x_{a}))$ in Section~\ref{s:sec.3}) so that the generator in the CGAN produces distinctive samples among the process states. Finally, Variant 3, which is the proposed method, adds the terms  ($\lVert \nabla D(\hat{x},y_{a}) \rVert_{2}$, and $D(x_{a},y_{m})$ in Section~\ref{s:sec.3}) that are related to stabilizing the learning process in the objective function of the discriminator in Variant 2.

Table~\ref{tab:gan_table2} shows the results of the ablation study. In addition to F-score, Precision, and Recall, Frechet Inception Distance \citep{dowson1982frechet} (FID) is used as an additional performance measure in the ablation study. FID is a metric used to evaluate the quality of images generated by GAN. Specifically, it is a metric between two multidimensional Gaussian distributions that are the distribution of neural network features from the actual and generated images, respectively. FID is computed from the mean and the covariance of the activation function of the network as follows:
\begin{equation*}
    \text{FID} = \lVert \mu_{a}-\mu_{g} \rVert^{2} + \text{Tr}(\Sigma_{a} + \Sigma_{g} -2(\Sigma_{a}\Sigma_{g})^{1/2}),
\end{equation*}
where $\mu_{a}$ and $\Sigma_{a}$ are the mean and standard deviation from actual images, respectively; likewise, $\mu_{g}$ and $\Sigma_{g}$ are those from generated images, respectively.
The smaller FID represents that the generated images follow the distribution of actual images resulting in better quality and diversity \citep{chen2021data}. In Table~\ref{tab:gan_table2}, the FID scores of each label are presented. 1500 images of each label from actual and generated images, respectively, are used to calculate the FID.
\vspace{-0.4cm}\begin{table}[!htb]
\renewcommand{\arraystretch}{1.2}
\centering
\caption{Performance evaluation in ablation studies.}\vspace{0.0cm}
\label{tab:gan_table2}
\resizebox{1.0\textwidth}{!}{%
\begin{tabular}{cccccccccc}
\hline \hline
\textbf{}  &   $G(z, y_{g})$ & \begin{tabular}[c]{@{}c@{}}\vspace{-0.0cm}$y_{a}\text{log}(C(x_{a})),$\\$y_{g}\text{log}(C(G(z,y_{g})))$\end{tabular} & \begin{tabular}[c]{@{}c@{}}\vspace{-0.0cm}$\lVert \nabla D(\hat{x},y_{a}) \rVert_{2},$\\$D(x_{a},y_{m})$\end{tabular}     & \textbf{ F-score} & \textbf{Precision} & \textbf{Recall} &
\textbf{\begin{tabular}[c]{@{}c@{}}\vspace{-0.0cm}FID\\T-shirt\end{tabular}}& \textbf{\begin{tabular}[c]{@{}c@{}}\vspace{-0.0cm}FID\\Pullover\end{tabular}}&\textbf{\begin{tabular}[c]{@{}c@{}}\vspace{-0.0cm}FID\\Dress\end{tabular}}\\
\cline{2-10}
Baseline & \xmark   &         \xmark     &  \xmark & 0.783            & 0.785              & 0.873           & NA            & NA            & NA            \\
Variant1 &  \checkmark     &    \xmark          &            \xmark & 0.791            & 0.789              & 0.879           & 203.9            & 239.9            & 226.1            \\
Variant2 &   \checkmark     &   \checkmark           &               \xmark & 0.806            & 0.797              & 0.886           & 192.3            & 234.9            & 227.9            \\
\begin{tabular}[c]{@{}c@{}}\vspace{0.0cm}\textbf{Variant3}\\ \vspace{+0.0cm}\textbf{(Proposed)}\end{tabular}  & \checkmark     &       \checkmark       &          \checkmark & \textbf{0.830}            & \textbf{0.814}              & \textbf{0.900}           &  \textbf{109.3}            & \textbf{142.9}            & \textbf{184.3}      \\
\hline \hline
\end{tabular}
 }
\end{table}\vspace{-0.5cm}

\begin{figure}[!htp]
    \centering
    {\includegraphics[width=0.7\textwidth]{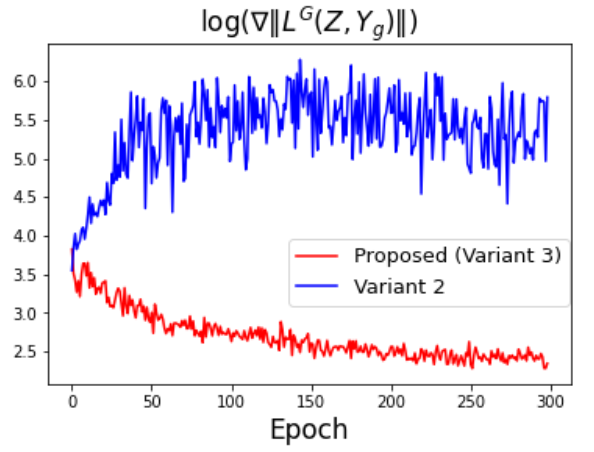}}\vspace{-0.0cm}
    \caption{Comparison of a log scale norm for the gradient  of the generator.}
    \label{fig:gan_gradient}
\end{figure}

The performance of all the measures of Variant 1 improves compared to the baseline. It shows the effectiveness of GAN for supplementing imbalanced training data. In Variant 2, an improvement in the performance of most of the measures is achieved compared to the first ablation. This implies that joint training of GAN and classifier guides the generator to generate samples beneficial to the classifier, that is, both realistic and state-distinguishable samples. It consequently improves the performance of the classifier. 
Finally, when both terms stabilizing the learning process of GAN are added, which is the proposed method (Variant 3), the performances are significantly improved since stable training enables the generator to generate more realistic and state-distinctive samples. Specifically, Fig.~\ref{fig:gan_gradient} compares the log scale norm for the gradient of the generator ($\log(\nabla \lVert L^{G}(Z,Y_{g})\rVert$) between Variant 2 and the proposed method. The results show the gradient of the proposed method decreases with the small variance, while the gradient of Variant 2 explodes with a large variance, resulting in unstable learning and low-quality generated samples  \citep{arjovsky2017towards}.   
\vspace{-0.6cm}\vspace{-0.6cm}

\begin{figure}[!htb]
\centering
\subfloat[]{%
\resizebox*{0.4\textwidth}{!}{\includegraphics{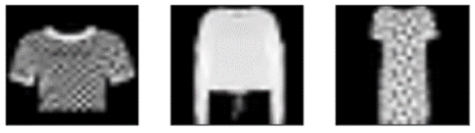}} \label{subfig:gan_3a}}\hspace{0pt}
\subfloat[]{%
\resizebox*{0.4\textwidth}{!}{\includegraphics{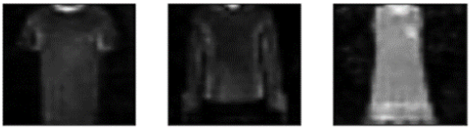}}\label{subfig:gan_3b}}\hspace{0pt}
\subfloat[]{%
\resizebox*{0.4\textwidth}{!}{\includegraphics{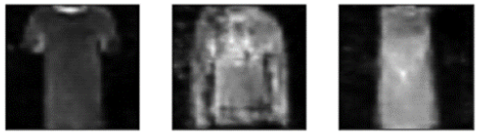}}\label{subfig:gan_3c}}\hspace{0pt} 
\subfloat[]{%
\resizebox*{0.4\textwidth}{!}{\includegraphics{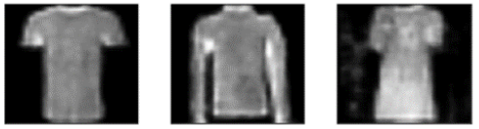}}\label{subfig:gan_3d}}\hspace{0pt} \\\vspace{-0.0cm}
\caption{Samples of T-shirt, Pullover, and Dress from (a) actual images, (b) generated images from Variant 1, (c) generated images from Variant 2, and (d) generated images from Variant 3 (Proposed method).}
 \label{fig:gan_3}
\end{figure}

Fig.~\ref{fig:gan_3} shows the quality of generated images in each step of the ablation study. Fig.~\ref{fig:gan_3} (a) represents the actual images from each label. Figs.~\ref{fig:gan_3} (b), (c), and (d) illustrate the generated images of each label from Variant 1, 2, and the proposed method, respectively. Like the results of the FID score in Table~\ref{tab:gan_table2}, the quality of generated samples is improved as each ablation is added.
\vspace{-0.5cm}\vspace{-0.7cm}

\begin{figure}[!htb]
\centering
\subfloat[]{%
\resizebox*{0.40\textwidth}{!}{\includegraphics{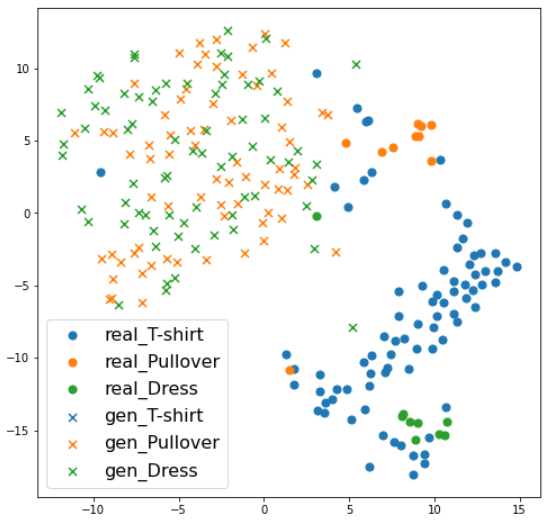}} \label{subfig:gan_4a}}\hspace{0pt}
\subfloat[]{%
\resizebox*{0.40\textwidth}{!}{\includegraphics{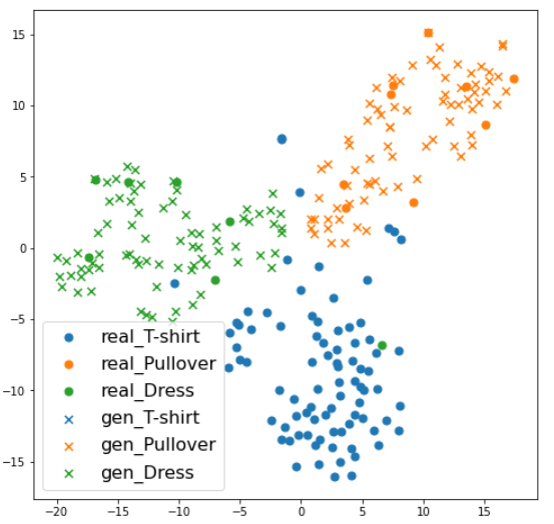}}\label{subfig:gan_4b}}\hspace{0pt}\\
\caption{t-SNE of the feature from the intermediate layer of the classifier from the proposed method in epochs (a) 0 and (b) 300.}\vspace{-0.5cm}
 \label{fig:gan_4}
\end{figure}

Beyond the image quality and FID score, Fig.~\ref{fig:gan_4} shows the effectiveness of the generated samples from the proposed method based on the comparison between the feature of generated and actual samples. Specifically, Fig.~\ref{fig:gan_4} represents the t-distributed Stochastic Neighbourhood Embedding (t-SNE) of the feature from the intermediate layer of the classifier in the proposed method. t-SNE is a tool for visualizing high-dimensional data \citep{dimitriadis2018t}. 
\vspace{-0.5cm}
It is a nonlinear dimensionality reduction technique suitable for incorporating high-dimensional data into lower-dimensional data (2-D or 3-D) for visualization. `$\bullet$' and `$\times$' in Fig.~\ref{fig:gan_4} denote the feature in the intermediate layer of classifiers from actual and the generated samples in the balanced training batch, respectively. For a balanced training batch, the minority class has many `$\times$' in each batch. Fig.~\ref{fig:gan_4} (a) shows the distribution alignments of actual and generated samples are different at epoch 0.  Since the proposed method learns to generate realistic and distinctive samples for the classifier, Fig.~\ref{fig:gan_4} (b) shows that the features of generated samples (`$\times$') correctly follow those of actual samples (`$\bullet$') according to their respective labels at epoch 300. In addition, features from each label are clearly separated. The balanced training data with these properties let the proposed method achieve high classification performance.

\vspace{-0.2cm}\subsection{Open-Source Data Case Studies} \label{s:sec4.2}
The open-source image data set is used for the comparative studies in this section. MNIST fashion \citep{xiao2017fashion} and CIFAR-10 \citep{recht2018cifar}, which are widely used for the evaluation of image-based classifiers, are selected. In MNIST fashion data, five labels related to upper clothes are selected for analysis. This step provides high similarity among the classes to make a challenging problem \citep{choi2021imbalanced}. For CIFAR-10 data, Airplane, Automobile, and Ship images are selected for the same reason as the MNIST fashion data. In each label of MNIST fashion and CIFAR-10 data, 1000 and 1500 images are collected, respectively. Then, imbalanced training data is provided in Table~\ref{tab:gan_table4}. The balanced ratios are determined as 0.10 and 0.30 for  MNIST fashion and CIFAR-10 data, respectively. This is because the balanced ratios with less than these values provide poor performance that is meaningless for the comparison. The remaining data sets are used as testing data.
\vspace{-0.6cm}\begin{table}[!htp]
\renewcommand{\arraystretch}{1.2}
\centering
\caption{Imbalanced training data samples in open-source data case studies.}
\label{tab:gan_table4}
\resizebox*{1.0\textwidth}{!}{%
\begin{tabular}{cccccc}
\hline \hline
 \textbf{\begin{tabular}[c]{@{}c@{}}\vspace{-0.0cm}Data\\Set\end{tabular}}        &  \textbf{\begin{tabular}[c]{@{}c@{}}\vspace{-0.0cm}Majority\\Class\end{tabular}} & \textbf{\begin{tabular}[c]{@{}c@{}}\vspace{-0.0cm}Minority\\Class\end{tabular}}                     & \textbf{\begin{tabular}[c]{@{}c@{}}\vspace{-0.0cm}Balanced\\Ratio\end{tabular}}  & \textbf{\begin{tabular}[c]{@{}c@{}}\vspace{-0.0cm}Majority Class\\Training Samples\end{tabular}}& \textbf{\begin{tabular}[c]{@{}c@{}}\vspace{-0.0cm}Minority Class\\Training Samples\end{tabular}} \\  \hline
MNIST fashion & T-shirt  & Pullover, Dress, Coat, Shirt & 0.10  & 800                                                             & 80                                                                \\
CIFAR-10        & Airplane & Automobile, Ship             & 0.30   & 1000                                                            & 300         \\     
\hline \hline                               
\end{tabular}
 }
\end{table}\vspace{-0.4cm}

Table~\ref{tab:gan_table5} shows the performance evaluation of the proposed and benchmark methods in two open-source data sets. The proposed method achieves the best performance in most measures in both data sets. Compared to a baseline using imbalanced training data to train the classifier, the sampling-based method such as SMOTE \citep{chawla2002smote}, B-SMOTE \citep{han2005borderline}, and ADASYN \citep{he2008adasyn} generally achieves similar or worse performance. 
\vspace{-0.5cm}\begin{table}[!htp]
\renewcommand{\arraystretch}{1.2}
\centering
\caption{Performance evaluation in open-source data case studies.}
\label{tab:gan_table5}
\resizebox*{0.9\textwidth}{!}{%
\begin{tabular}{lccc|ccc}
\hline \hline
\textbf{}            & \multicolumn{3}{c}{\textbf{MNIST fashion}}                   & \multicolumn{3}{c}{\textbf{CIFAR-10}}                    \\
\cline{2-7}
\textbf{}            & \textbf{F-score} & \textbf{Precision} & \textbf{Recall} & \textbf{F-score} & \textbf{Precision} & \textbf{Recall} \\
\cline{2-7}
\textbf{Baseline}    & 0.621            & 0.650              & 0.696           & 0.617            & 0.661              & 0.670           \\
\textbf{SMOTE}       & 0.612            & 0.645              & 0.687           &       0.618           &        0.660            &          0.669       \\
\textbf{B-SMOTE}     & 0.610            & 0.643              & 0.683           & 0.642            & 0.642              & 0.631           \\
\textbf{ADASYN}      & 0.604            & 0.639              & 0.679           & 0.559           & 0.638              & 0.627           \\
\textbf{CDRAGAN}     & 0.632            &     0.657               &     0.703            &     0.597             &   0.648                  &           0.655      \\
\textbf{BAGAN-GP}   & 0.634            & 0.661              & 0.708           & 0.634            & \textbf{0.664}              & 0.680           \\
\textbf{Cooperative GAN} & 0.619            & 0.646              & 0.693           & 0.619            & 0.628              & 0.656           \\
\textbf{Proposed}     & \textbf{0.642}            &  \textbf{0.664}              & \textbf{0.713}           &  \textbf{0.652}            & 0.655              & \textbf{0.684} \\   \hline \hline      
\end{tabular}
}\vspace{-0.0cm}
\end{table}
Since both case studies have a small number of minority data, the sampling-based method has limitations in generating various data that can cover the testing data. In contrast, the GAN-based techniques usually achieve better performance than sampling-based methods since their generators learn the actual distribution and produce various training data for the classifier. Especially, the generator from the proposed method provides more diverse and better quality images by jointly optimizing the classifier, resulting in improvements in classification results.

\vspace{-0.2cm}\subsection{Polymer Additive Manufacturing Process Data Case Studies} \label{s:sec4.3}
\begin{figure}[!htb]
\centering
\subfloat[]{%
\resizebox*{0.3\textwidth}{!}{\includegraphics{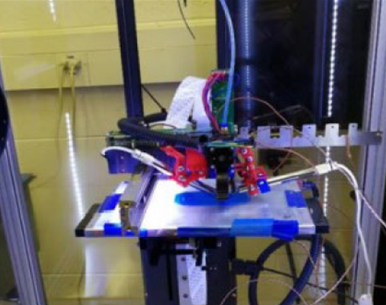}} \label{subfig:gan_5a}}\hspace{0pt}
\subfloat[]{%
\resizebox*{0.3\textwidth}{!}{\includegraphics{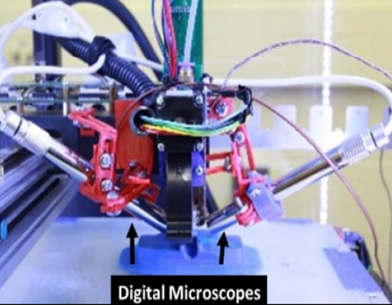}}\label{subfig:gan_5b}}\hspace{0pt}
\subfloat[]{%
\resizebox*{0.3\textwidth}{!}{\includegraphics{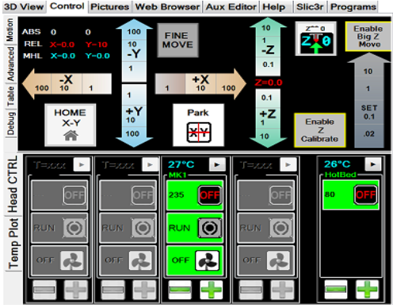}}\label{subfig:gan_5c}} \\\vspace{-0.0cm}
\caption{(a) Front view of Hyrel system 30M; (b) Digital Microscope Camera; (c) Software Controller.}
 \label{fig:gan_5}
\end{figure}

\begin{figure}[!htb]
\centering
\subfloat[]{%
\resizebox*{0.3\textwidth}{!}{\includegraphics{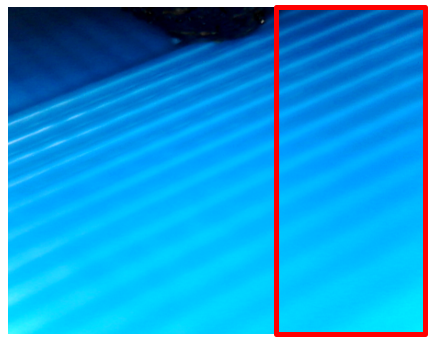}} \label{subfig:gan_6a}}\hspace{0pt}
\subfloat[]{%
\resizebox*{0.3\textwidth}{!}{\includegraphics{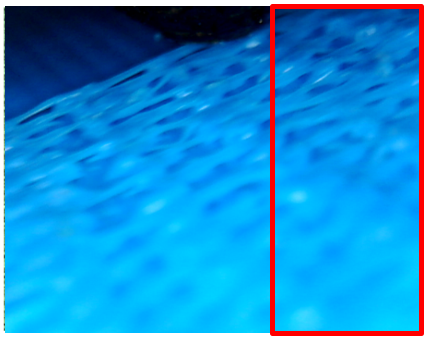}}\label{subfig:gan_6b}}\hspace{0pt}
\subfloat[]{%
\resizebox*{0.3\textwidth}{!}{\includegraphics{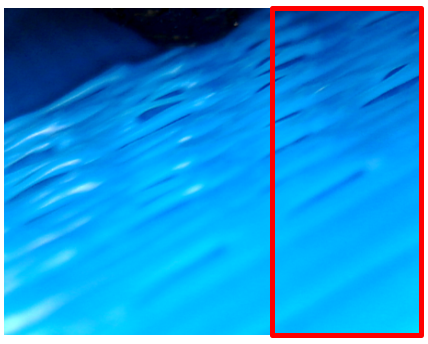}}\label{subfig:gan_6c}} \\\vspace{-0.0cm}
\caption{(a) Normal; (b) Under-fill caused by feed rate; (c) Under-fill caused by a cooling fan. The red rectangle represents the regions of interest in each image.}\vspace{-0.0cm}
 \label{fig:gan_6}
\end{figure}

 A Hyrel System 30M 3-D printer equipped with a 0.5 mm extruder nozzle is used for this case study. Fig.~\ref{fig:gan_5} (a) shows a front view of the printer. Acrylonitrile butadiene styrene (ABS), with a diameter of 1.75 mm, is used as a filament for printing. This study uses a digital microscope camera for high-quality image acquisition at a sampling frequency of 1 Hz. The camera is mounted near the extruder to collect images of the surface that is being printed (Fig.~\ref{fig:gan_5} (b)). 
\vspace{-0.5cm}\vspace{-0.5cm}
Based on the design of experiments in \cite{liu2019image}, the surface images for normal state, under-fill caused by feed rate and under-fill caused by the cooling fan, as shown in Fig.~\ref{fig:gan_6}, can be obtained by setting up the specific machine parameters in the software controller (Fig.~\ref{fig:gan_5} (c)).
\vspace{-0.0cm}\vspace{-0.0cm}
 Two process states which are the under-fill caused by feed rate and under-fill caused by the cooling fan, are the abnormal states that cause quality deterioration in the FFF process \citep{liu2019image,shen2020clustered}.
 To increase the size of image samples, the region of interest is one-third of each image in Fig.~\ref{fig:gan_6} with the size of 460 $\times$ 213 $\times$ 3.
 Therefore, each image collected from a microscope provides three image samples. In total, there are 915 samples from the normal process state and 591 and 459 samples from under-fill caused by feed rate and cooling fan, respectively.
This section provides case studies with various balanced ratios between normal and abnormal states of the FFF process to see the performance change according to ratio changes. The balanced ratios of training data are provided in Table~\ref{tab:gan_table6}, where the minority classes are denoted as the cause of abnormal states in the process. Specifically, the balanced ratios provide the performances of F-score, Precision, and Recall that are applicable in practice are utilized. The remaining images in each process state are used as testing data.
\vspace{-0.6cm}\begin{table}[!htp]
\renewcommand{\arraystretch}{1.2}
\centering
\caption{Imbalanced training data samples in Polymer AM case studies.}\vspace{-0.0cm} 
\label{tab:gan_table6}
\resizebox*{1.0\textwidth}{!}{%
\begin{tabular}{ccccc}
\hline \hline
 \textbf{\begin{tabular}[c]{@{}c@{}}\vspace{-0.0cm}Majority\\Class\end{tabular}} & \textbf{\begin{tabular}[c]{@{}c@{}}\vspace{-0.0cm}Minority\\Class\end{tabular}}                     & \textbf{\begin{tabular}[c]{@{}c@{}}\vspace{-0.0cm}Balanced\\Ratio\end{tabular}}  & \textbf{\begin{tabular}[c]{@{}c@{}}\vspace{-0.0cm}Majority Class\\Training Samples\end{tabular}}& \textbf{\begin{tabular}[c]{@{}c@{}}\vspace{-0.0cm}Minority Class\\Training Samples\end{tabular}} \\  \hline
 Normal  & Under-feed, Under-fan & 0.10  & 800                                                            & 80                                                                \\
 Normal  & Under-feed, Under-fan             & 0.15   & 800                                                           & 120         \\  Normal  & Under-feed, Under-fan          & 0.20   & 800                                                            & 160         \\     
\hline \hline                               
\end{tabular}
 }\vspace{-0.3cm}
\end{table}

\begin{figure*}[b]
    \centering
    \resizebox*{1.0\textwidth}{!}{\includegraphics{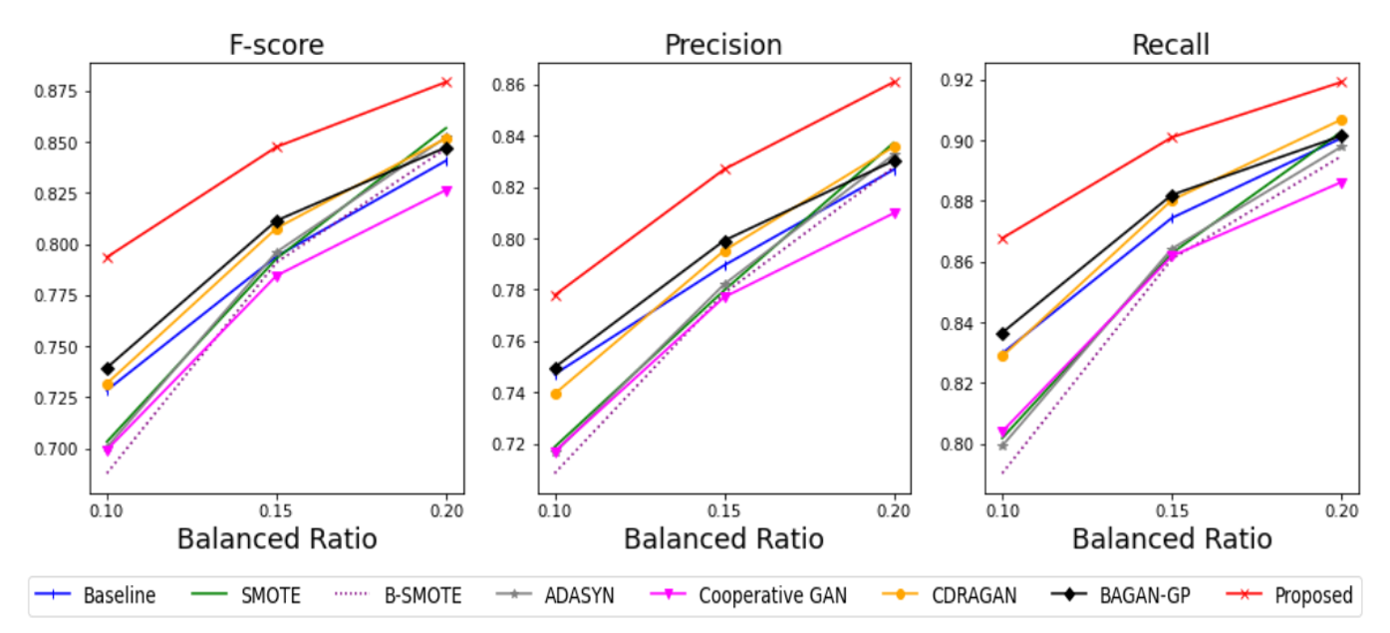}}\vspace{-0.0cm}
    \caption{F-scores, precisions, and Recalls in FFF process case studies under different balanced ratios.}
    \label{fig:gan_7}
\end{figure*}

Fig.~\ref{fig:gan_7} shows the performance evaluation of the proposed and benchmark methods in various balanced ratios. 
\vspace{-0.0cm}
The performances of all the methods are improved when the number of training samples in minority classes increases (i.e., the balanced ratio increases). This is because a large number of samples provides more information for the generator to learn the actual distribution. In every balanced ratio, the proposed method achieves the best performance in all the measures, which shows the effectiveness of the realistic and state-distinguishable generated samples in the AM process. Specifically, the proposed method improved 3\%$\sim$13\%, 
3\%$\sim$10\%, and 2\%$\sim$10\% of the performance of the benchmark methods regarding their F-score, Precision, and Recall, respectively. In addition, the proposed method performs better than cooperative GAN, which also jointly optimizes GAN and classifier but follows the basic objective function of the discriminator in GAN. This demonstrates the effectiveness of stable learning through regularizing the gradient of the discriminator and an additional task provided to the discriminator in the proposed method.
Sampling-based methods show worse performance than GAN-based methods in general since the methods only consider the local information resulting in limited diverse generated images \citep{douzas2018effective}.

\begin{figure}[b]
\centering
\subfloat[]{%
\resizebox*{0.40\textwidth}{!}{\includegraphics{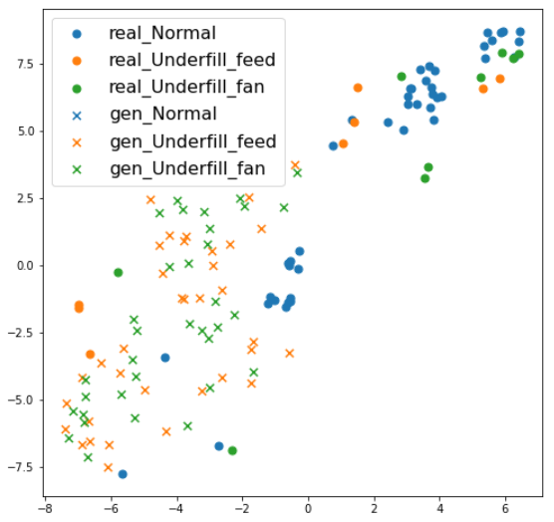}} \label{subfig:gan_8a}}\hspace{0pt}
\subfloat[]{%
\resizebox*{0.40\textwidth}{!}{\includegraphics{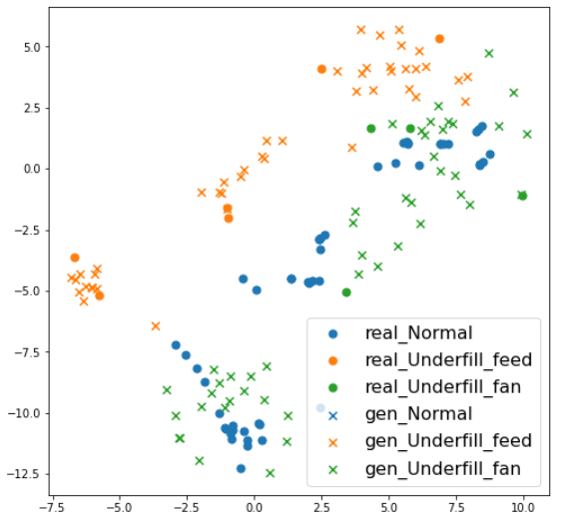}}\label{subfig:gan_8b}}\hspace{0pt}\\
\caption{t-SNE of the feature from the intermediate layer of classifiers from the proposed method in epochs (a) 0 and (b) 300 in the FFF process when the balanced ratio is 0.2.}\vspace{-0.0cm}
 \label{fig:gan_8}
\end{figure}

\begin{figure}[!htp]
\centering
\subfloat[]{%
\resizebox*{0.4\textwidth}{!}{\includegraphics{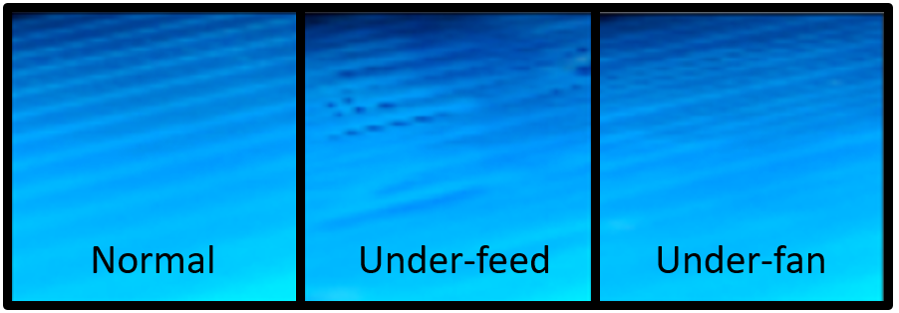}} \label{subfig:gan_9a}}\hspace{0.15in}
\subfloat[]{%
\resizebox*{0.4\textwidth}{!}{\includegraphics{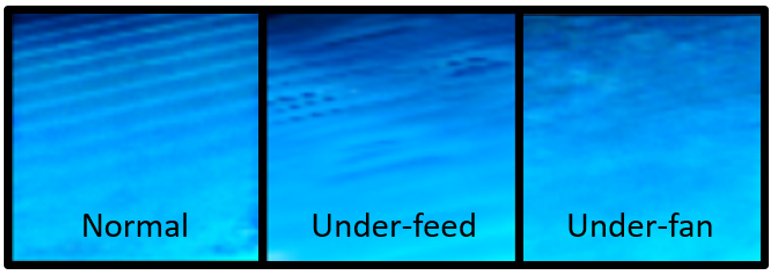}}\label{subfig:gan_9b}}
\vspace{-0.0cm}\caption{Samples of each process state in the FFF process from (a) actual; (b) generator when the
balanced ratio is 0.2.}
 \label{fig:gan_9}
\end{figure}

Fig.~\ref{fig:gan_8} shows the t-SNE of the feature from the intermediate layer of classifiers from the proposed method in epochs 0 and 300 when the balanced ratio is 0.20. Like Fig.~\ref{fig:gan_4}, `$\bullet$' and `$\times$' represent features of actual and generated samples, respectively. The colors differentiate each process state in the FFF process. 
\vspace{-0.0cm}
To make a balanced training data, two abnormal states in the FFF process have adequate generated samples (`$\times$') than actual samples (`$\bullet$') in each batch. Compared to epoch 0 (Fig.~\ref{fig:gan_8} (a)), features in epoch 300 (Fig.~\ref{fig:gan_8} (b)) show that the features from generated samples (`$\times$') of abnormal states of the FFF process follow those of actual samples (`$\bullet$') correctly to each process state. 
Based on these balanced training data in the FFF process, the classifier achieves the best prediction results compared to benchmark methods. Fig.~\ref{fig:gan_9} shows the samples of actual and generated images denoted as `$\bullet$' and `$\times$' in Fig.~\ref{fig:gan_8} (b), respectively. The generated images (Fig.~\ref{fig:gan_9} (b)) are realistic and distinguishable among process states based on both adversarial and cooperative learning in the proposed method. 

\vspace{-0.4cm}\vspace{-0.0cm}
\subsection{Metal Additive Manufacturing Process Data Case Studies} \label{s:sec4.4}
\begin{figure}[!htp]
\centering
\subfloat[]{%
\resizebox*{0.3\textwidth}{!}{\includegraphics{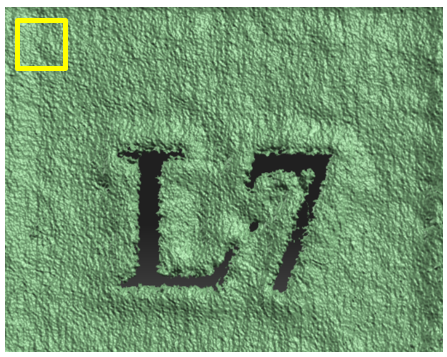}} \label{subfig:gan_10a}}\hspace{0pt}
\subfloat[]{%
\resizebox*{0.3\textwidth}{!}{\includegraphics{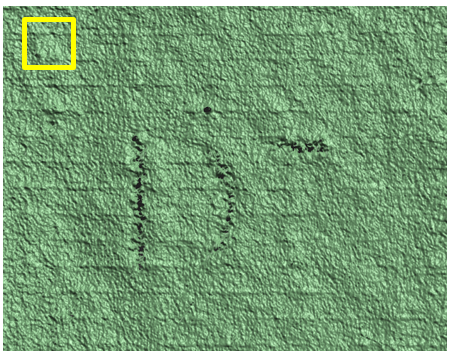}}\label{subfig:gan_10b}}\hspace{0pt}
\subfloat[]{%
\resizebox*{0.3\textwidth}{!}{\includegraphics{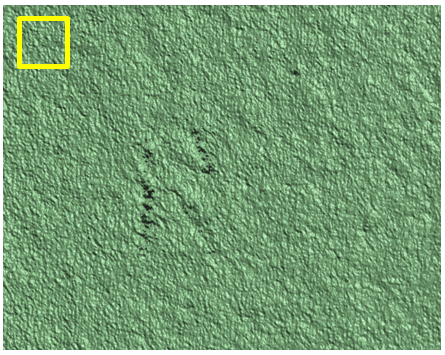}}\label{subfig:gan_10c}} \\
\caption{(a) Raster. (b) Dehoff. (c) Random; L7, D7, and R are the markers to identify samples using different scan strategies. The yellow square indicates the region of interest.}\vspace{-0.0cm}
 \label{fig:gan_10}
\end{figure}

The machine ARCAMQ10 plus is utilized to print samples from the EBM process using Ti-6A1-4V powder. The dimensions of the printed sample are 15 mm $\times$ 15 mm $\times$ 25 mm. In the EBM process, there exist three different scan strategies, i.e., Dehoff, Raster, and Random \citep{kirka2017strategy}. The different scan strategies provide different surface patterns for the printed samples. After printing three different samples with each scan strategy, a 3-D scanner captures detailed 3-D information about the top surface (15 mm $\times$ 15 mm) quality \citep{wang2021development}. The images shown in Fig.~\ref{fig:gan_10} are the surface patterns of the Raster, Dehoff, and Random, respectively. 
\vspace{-0.5cm}\vspace{-0.5cm}

The objective of classification in this case study is to identify the scan strategy from the surface images. For each image in Fig.~\ref{fig:gan_10}, the size is 824 $\times$ 1118 $\times$ 3. To obtain training samples, 322 images with sizes of 120 by 120 are collected from the upper part of each image  (300 $\times$ 1118 $\times$ 3) since the bottom parts of the surface image with letters and numbers have many defects, such as porosity. Specifically, the collected images are highly overlapped in the horizontal directions (114 pixels) to have plenty of samples. Since the Raster scan strategy is commonly used and similar to the common AM bi-directional path, the strategy is considered a majority class \citep{saville2021texture, nandwana2020influence}. Therefore, the scan strategies with Dehoff and Random are determined as minority classes in these case studies. From the 322 images from each scan strategy, the various balanced ratios of training data are designed as in Table~\ref{tab:gan_table7}.
As with polymer AM case studies in Section~\ref{s:sec4.3}, the balanced ratios that provide the reasonable F-score, Precision, and Recall applicable in practice are utilized in case studies. The remaining images in each scan strategy are used as testing data.
\vspace{-0.5cm}\begin{table}[!htp]
\renewcommand{\arraystretch}{1.2}
\centering
\caption{Imbalanced training data samples in Metal AM case studies.}
\label{tab:gan_table7}
\resizebox*{1.0\textwidth}{!}{%
\begin{tabular}{ccccc}
\hline \hline
\textbf{\begin{tabular}[c]{@{}c@{}}\vspace{-0.0cm}Majority\\Class\end{tabular}} & \textbf{\begin{tabular}[c]{@{}c@{}}\vspace{-0.0cm}Minority\\Class\end{tabular}}                     & \textbf{\begin{tabular}[c]{@{}c@{}}\vspace{-0.0cm}Balanced\\Ratio\end{tabular}}  & \textbf{\begin{tabular}[c]{@{}c@{}}\vspace{-0.0cm}Majority Class\\Training Samples\end{tabular}}& \textbf{\begin{tabular}[c]{@{}c@{}}\vspace{-0.0cm}Minority Class\\Training Samples\end{tabular}} \\  \hline
Raster  & Dehoff, Random & 0.3  & 150                                                             & 45                                                               \\
Raster  & Dehoff, Random             & 0.4   & 150                                                           & 50         \\  
Raster  & Dehoff, Random          & 0.5   & 150                                                            & 75         \\     
\hline \hline                               
\end{tabular}
}
\end{table}\vspace{-0.2cm}

\begin{figure*}[b]
    \centering
    \resizebox*{1.0\textwidth}{!}{\includegraphics{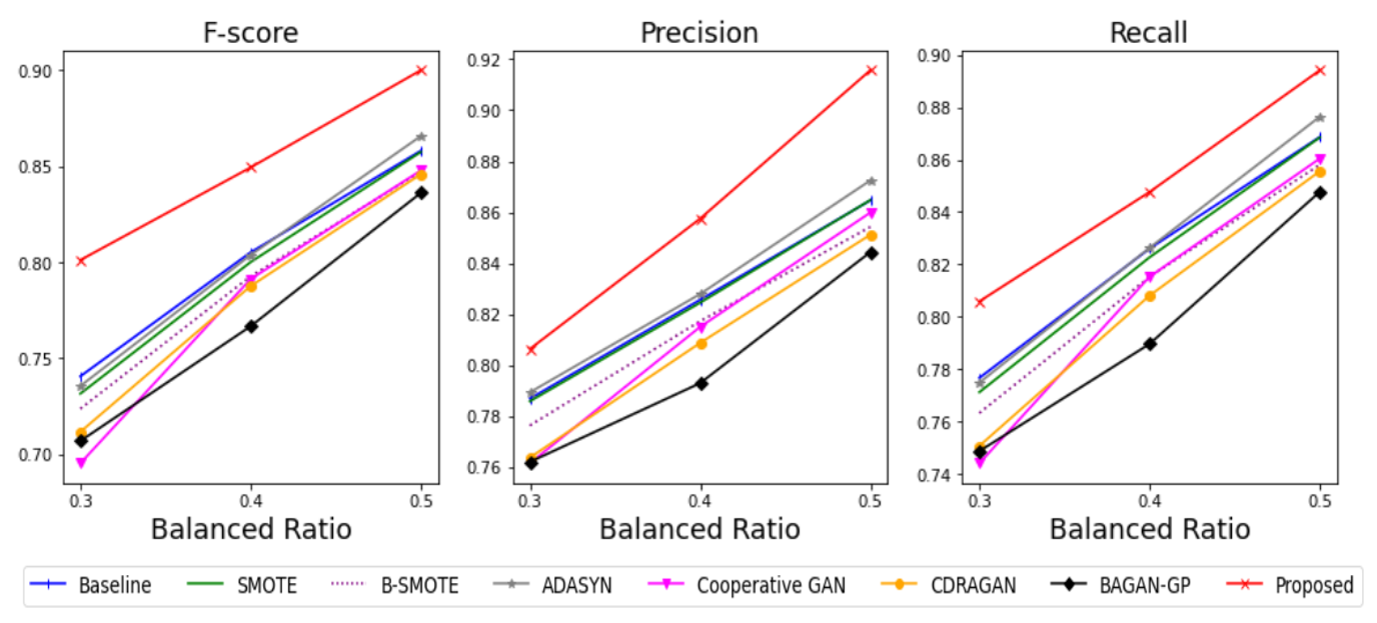}}\vspace{-0.0cm}
    \caption{F-scores, Precisions, and Recalls in EBM process case studies under different balanced ratios.}
    \label{fig:gan_11}
\end{figure*}

\begin{figure}[!htb]
\centering
\subfloat[]{%
\resizebox*{0.40\textwidth}{!}{\includegraphics{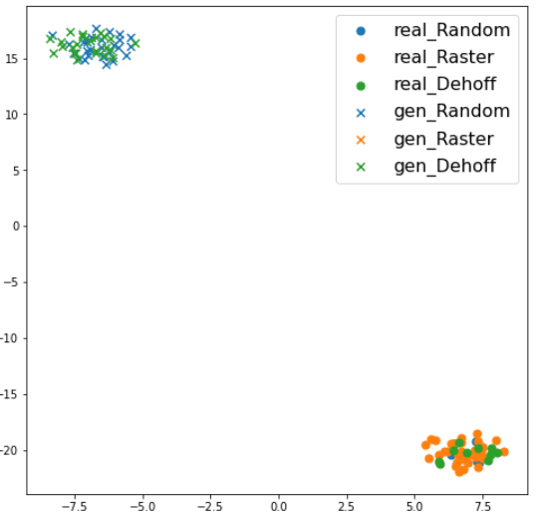}} \label{subfig:gan_12a}}\hspace{0pt}
\subfloat[]{%
\resizebox*{0.40\textwidth}{!}{\includegraphics{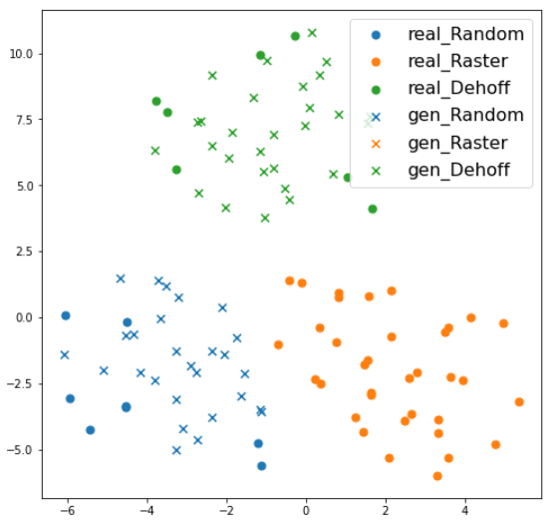}}\label{subfig:gan_12b}}\hspace{0pt}\\\vspace{-0.0cm}
\caption{t-SNE of the feature from the intermediate layer of classifiers from the proposed method in epochs (a) 0 and (b) 300 in the EBM process when the balanced ratio is 0.3.}\vspace{-0.0cm}
 \label{fig:gan_12}
\end{figure}

\begin{figure}[!htb]
\centering
\subfloat[]{%
\resizebox*{0.4\textwidth}{!}{\includegraphics{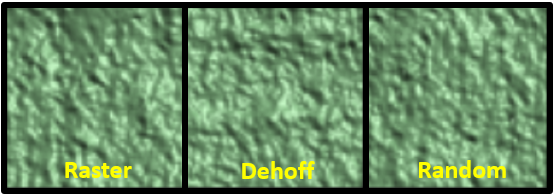}} \label{subfig:gan_13a}}\hspace{0.0in}
\subfloat[]{%
\resizebox*{0.4\textwidth}{!}{\includegraphics{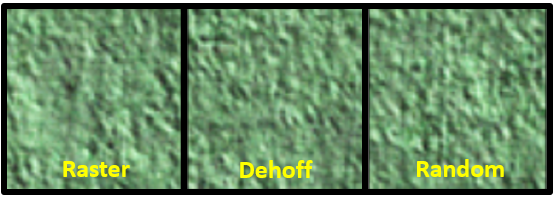}}\label{subfig:gan_13b}}\hspace{-0.1pt}\\\vspace{-0.0in}
\caption{Samples of each scan strategy in the EBM process from (a) actual; (b) generator when balanced ratio is 0.3.}
 \label{fig:gan_13}
\end{figure}

Fig.~\ref{fig:gan_11} shows the performance evaluation of the proposed and benchmark methods in the EBM process. In this case study, most sampling-based methods perform better than the baseline, but the GAN-based methods usually represent worse results than the baseline. This might be caused by the highly overlapped actual samples. Compared to the polymer AM case study, the number of actual images is small and highly overlapped since we only have a single image of the top surface. 
\vspace{-0.0cm}
Therefore, it does not provide enough information for the GAN-based methods to learn the actual distribution. Relatively, the sampling-based methods show better performance because the actual images are highly overlapped with each other. Still, the proposed method achieves the best results compared to benchmark methods by generating balanced training samples that are both realistic and scan strategy-distinguishable. Specifically, the F-score, Precision, and Recall of the proposed method were improved 4\%$\sim$15\%, 2\%$\sim$8\%, and 3\%$\sim$8\% over those of benchmark methods, respectively.

t-SNE results in Fig.~\ref{fig:gan_12} show similar results to the FFF process in Fig.~\ref{fig:gan_8}. In epoch 300, the features from generated samples follow those of corresponding scan strategies from actual samples. Fig.~\ref{fig:gan_13} shows the actual and generated samples of all the scan strategies from the proposed method denoted as `$\bullet$' and `$\times$' in Fig.~\ref{fig:gan_12} (b), respectively. Compared to the actual images in Fig.~\ref{fig:gan_13} (a), the generated images in the EBM process (Fig.~\ref{fig:gan_13} (b)) are realistic. In addition, the generated images are distinguishable according to each scan strategy, which is validated by the superior classification performance of the proposed method.
\vspace{-0.3cm}\vspace{-0.4cm}
\vspace{-0.4cm}
\section{Conclusions}\label{s:sec5}
This paper proposes a novel GAN-based data augmentation method to deal with the imbalanced training data issue in manufacturing processes. The method consists of three-player, namely, generator, discriminator, and classifier, jointly optimized. Through the adversarial learning between the generator and the discriminator, the generator generates realistic samples of abnormal states in the manufacturing process. At the same time, the cooperative learning between the generator and classifier enables the generator to generate the state-distinguishable samples of the manufacturing process. 
The generated samples are added to actual samples and provided as a balanced training batch for a classifier. In addition, the method regularizes the gradient of the discriminator and  provides an additional task to the discriminator compared to standard GAN. It prevents the gradient exploding of the generator resulting in a better quality of generated samples. The iterative learning process among these three players finally provides a classifier with high performance in classification results. The effectiveness of the proposed method is validated in both open-source and actual AM process data. Specifically, the method achieves the best performance compared to benchmark methods in the various balanced ratio of training data between normal and abnormal states in both the FFF and EBM processes.

\bmhead{Acknowledgments}
The research reported in this publication was supported by the Office of Naval Research under award N00014-18–1-2794, and the Department of Defense under award N00014-19–1-2728.


\bibliography{sn-bibliography}

\begin{appendices}
\newpage\section{Structure and Hyperparameters in the Methods}\label{app:app1}
In this section, the detailed structure and hyperparameters of the proposed and benchmark methods are provided. Table~\ref{tab:gan_table8} provides the hyperparameters of each method.
\vspace{-0.0cm}\begin{table}[!htb]
\renewcommand{\arraystretch}{1.2}
\centering
\caption{Hyperparameters of each method.}
\label{tab:gan_table8}
\resizebox*{0.9\textwidth}{!}{%
\begin{tabular}{ccc}
\hline \hline
Methods                             & Parameters        & Value \\ \cline{1-3}
SMOTE, ADASYN                        & Nearest K samples & 5     \\ \cline{1-3}
\multirow{2}{*}{B-SMOTE}                             & Type              & 1     \\ \cline{2-3} &  Nearest K samples & 5\\ \hline
\multirow{15}{*}{\begin{tabular}[c]{@{}c@{}}CDRAGAN\\BAGAN-GP\\Cooperative GAN\\Proposed\end{tabular}}
& Number of iterations & 300    \\ \cline{2-3} 
& \begin{tabular}[c]{@{}c@{}}Training ratio\\between generator and discriminator\end{tabular}  & 10\\ \cline{2-3} 
& Optimizer & Adam\\ \cline{2-3} 
& Learning rate & 0.0002    \\ \cline{2-3} 
& Momentum & $\beta_{1}=0.5, \beta_{2}=0.9$    \\ \cline{2-3} 
& Hidden layers (Discriminator)  & 
\begin{tabular}[c]{@{}c@{}}4 blocks of\\ \big[Conv2D, LeakyRelu\big]\end{tabular} \\ \cline{2-3}
& Hidden layers (Generator)  &

\begin{tabular}[c]{@{}c@{}}4 blocks of\\\big[Conv2D-Transpose, LeakyRelu, \\BatchNormalization\big] \end{tabular} \\ \cline{2-3}
& 
\begin{tabular}[c]{@{}c@{}}Number of Kernels in each block\\(Discriminator) \end{tabular}
 & (64,128,128,256)  \\ \cline{2-3}
& 

\begin{tabular}[c]{@{}c@{}}Number of Kernels in each block\\(Generator) \end{tabular} & (128,128,64,Number of channel)  \\ \cline{2-3}
& Kernel sizes & (4,4)  \\ \cline{2-3}
& Strides & (2,2)  \\ \cline{2-3}
& Padding & Same  \\ \cline{2-3}
& Activation functions & LeakyRelu, Tanh  \\ \cline{2-3}
& Kernel initializer & Random Normal(sd=0.02) \\ \cline{2-3}
& Slope of Leaky Relu & 0.2 \\ \cline{1-3}
\begin{tabular}[c]{@{}c@{}}CDRAGAN, BAGAN-GP\\Proposed\end{tabular}                  & Gradient Penalty Coefficient  & 10  \\ \cline{1-3}
Cooperative GAN                      & Range of scheduling parameter  & [0.1,0.9]  \\\cline{1-3} 
BAGAN-GP, Proposed                  & Epochs in pre-training  & 300  \\ 
\hline \hline   
\end{tabular}
}
\end{table}

The gradient penalty coefficient is determined as 10 as suggested by \cite{kodali2017convergence, huang2021enhanced}. The scheduling parameter in
the Cooperative GAN, is searched within a specific range ([0.1, 0.9]) following the guidelines
provided in the literature \citep{choi2021imbalanced} and selected with the values that showed the best validation performance. A batch size ($m$ in  Algorithm~\ref{alg:alg1}) varies along the number of actual samples in each case study since the proposed method is highly computationally expensive. Specifically, batch sizes are 100 in Section.~\ref{s:sec4.1}, 200 \& 100 in Section.~\ref{s:sec4.2}, 60 in Section.~\ref{s:sec4.3}, and 50 in Section.~\ref{s:sec4.4}.
Table~\ref{tab:gan_table9} shows the hyperparameters of the classifier in case studies. Convolutional Neural Network is used for the classifier. For a fair comparison, all the methods use the same classifier as described in Table~\ref{tab:gan_table9}.
\vspace{-0.0cm}\begin{table}[h]
\renewcommand{\arraystretch}{1.2}
\centering
\caption{Hyperparameters of the classifier.}
\label{tab:gan_table9}
\resizebox*{0.8\textwidth}{!}{%
\begin{tabular}{cc}
\hline \hline
Parameters                  & Value \\ \hline
Number of epochs         &   300    \\\hline
Optimizer               &  Adam    \\\hline
Learning rate               &  0.0002    \\\hline
Momentum               &  $\beta_{1}=0.5, \beta_{2}=0.9$    \\\hline
Hidden Layers                &   
\begin{tabular}[c]{@{}c@{}}4 blocks of\\ \big[Conv2D, LeakyRelu\big]\end{tabular}
  \\ \hline
Number kernels in each block & (32,32,128,256)      \\ \hline
Kernel sizes                 &  (4,4)     \\ \hline
Strides                      &  (2,2)     \\ \hline
Padding                      &  Same     \\ \hline
Activation functions         &   Leaky Relu, Softmax    \\ \hline
Kernel initializer           &  Random Normal (sd=0.02) \\ \hline
Slope of Leaky Relu          &   0.2  \\ 
\hline \hline
\end{tabular}
}
\end{table}

\end{appendices}
\end{document}